\documentclass[11pt]{article}

\usepackage[preprint]{acl}

\usepackage{times}
\usepackage{latexsym}
\usepackage{amsmath}
\usepackage{amssymb}
\usepackage{booktabs}
\usepackage{multirow}
\usepackage[T1]{fontenc}

\usepackage[utf8]{inputenc}

\usepackage{microtype}

\usepackage{inconsolata}

\usepackage{graphicx}

\title{Decision-Level Ordinal Modeling for Multimodal Essay Scoring with Large Language Models}

\author{Han Zhang, Jiamin Su, Li liu}

\begin{document}
\maketitle
\begin{abstract}
Automated essay scoring (AES) predicts multiple rubric-defined trait scores for each essay, where each trait follows an ordered discrete rating scale.
Most LLM-based AES methods cast scoring as autoregressive token generation and obtain the final score via decoding and parsing, making the decision implicit.
This formulation is particularly sensitive in multimodal AES, where the usefulness of visual inputs varies across essays and traits. To address these limitations, we propose \textbf{D}ecision-\textbf{L}evel \textbf{O}rdinal \textbf{M}odeling (\textbf{DLOM}), which makes scoring an explicit ordinal decision by reusing the LM head to extract score-wise logits on predefined score tokens, enabling direct optimization and analysis in the score space.
For multimodal AES, \textbf{DLOM-GF} introduces a gated fusion module that adaptively combines textual and multimodal score logits.
For text-only AES, \textbf{DLOM-DA} adds a distance-aware regularization term to better reflect ordinal distances.
Experiments on the multimodal EssayJudge dataset show that DLOM improves over a generation-based SFT baseline across scoring traits, and DLOM-GF yields further gains when modality relevance is heterogeneous.
On the text-only ASAP/ASAP++ benchmarks, DLOM remains effective without visual inputs, and DLOM-DA further improves performance and outperforms strong representative baselines.
\end{abstract}

\section{Introduction}
Automated Essay Scoring (AES) is a long-standing problem in natural language processing~\cite{14,15}, aiming to automatically assess the quality of student essays according to predefined scoring rubrics~\cite{13}, often across multiple scoring traits rather than a single holistic score~\cite{20,19,3}. Recent advances in large language models (LLMs) have significantly improved semantic understanding and instruction following, making them a natural choice for AES. As a result, a growing body of work has explored LLM-based essay scoring~\cite{20,16,17,25}, including extensions to multimodal settings where essays are accompanied by visual inputs such as charts or diagrams, which further complicates the scoring decision~\cite{1,2}.

\begin{figure}[t]
    \centering
    \includegraphics[width=\linewidth]{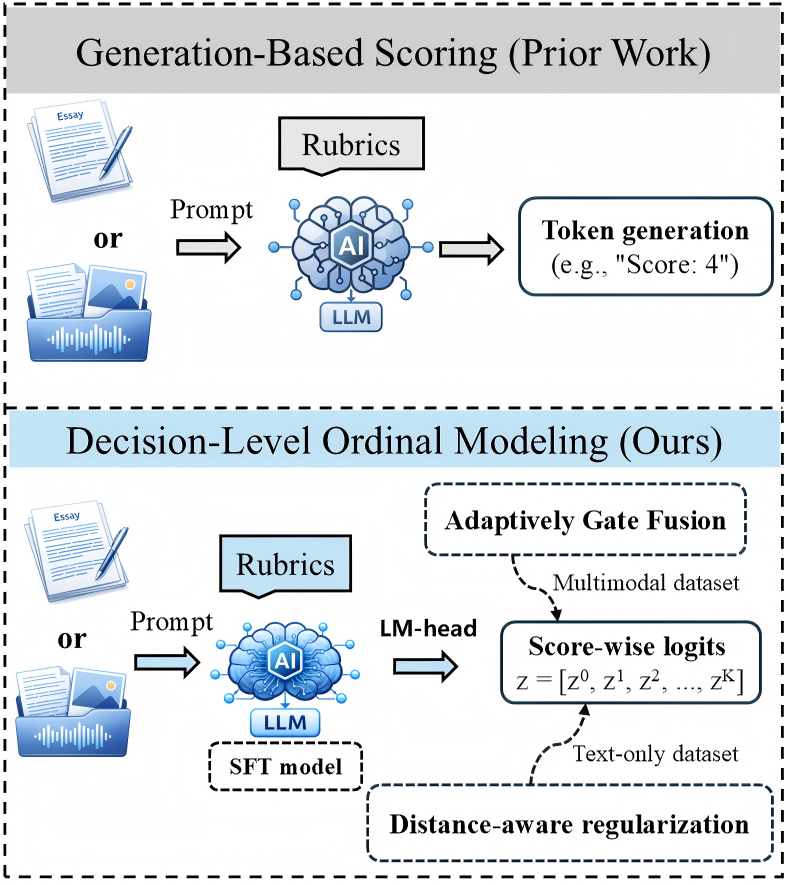}
   \caption{Comparison between generation-based scoring and decision-level ordinal modeling (DLOM).}
    \label{fig:introduction}
\end{figure} 

Most existing LLM-based AES approaches formulate scoring as a generation problem~\cite{3,4}, where the model is prompted to produce a score token or short textual response corresponding to a rubric-defined grade. This formulation is appealing due to its simplicity and compatibility with instruction-tuned LLMs~\cite{23,24,22}. However, generation-based scoring makes the scoring decision implicit, as the final score emerges as a byproduct of language generation~\cite{30}. Consequently, the decision process is difficult to control, analyze, or adapt, and is often sensitive to prompt design, decoding strategies, and tokenization artifacts—factors that are largely orthogonal to the scoring task itself~\cite{26}. 

In particular, AES is inherently ordinal: score categories follow a fixed and ordered structure, and errors between adjacent score levels are typically less severe than those between distant levels~\cite{29,27,28}. 
Motivated by this, we adopt a decision-based formulation of essay scoring to better match the ordinal nature of rubric-based assessment. 
We further argue that, especially with LLMs that provide strong semantic representations, separating semantic understanding from decision making offers a more appropriate and general modeling perspective for essay scoring.

In this work, we propose Decision-Level Ordinal Modeling (DLOM) for LLM-based essay scoring. 
Figure 1 illustrates the difference between generation-based scoring and our proposed decision-level ordinal modeling (DLOM). Building on a supervised fine-tuned (SFT) LLM, we cast score prediction as an explicit decision over a predefined ordered label set. The model extracts score-wise logits over the ordinal categories, enabling direct optimization and analysis at the decision stage while preserving the semantic representations learned during SFT. We further extend DLOM to multimodal scoring with a decision-level gated fusion module (\textbf{DLOM-GF}) that adaptively combines text-only and multimodal score logits.
For text-only benchmarks, we introduce a distance-aware regularization variant (\textbf{DLOM-DA}) to better respect ordinal distances between score levels.

Our main contributions are threefold:
\begin{itemize}
    \item \textbf{Decision-level ordinal modeling for LLM-based AES.}
    We cast scoring as an explicit ordinal decision by predicting score-wise logits over predefined score levels, enabling direct optimization and analysis in the score space.

    \item \textbf{DLOM extensions for multimodal and text-only scoring.}
    We introduce DLOM-GF for decision-level multimodal gated fusion and DLOM-DA for distance-aware regularization in text-only settings.

    \item \textbf{Comprehensive evaluation on multimodal and text-only benchmarks.}
    We demonstrate the effectiveness and generality of the proposed formulation on both multimodal and text-only essay scoring benchmarks, achieving consistent improvements over generation-based SFT baselines.
\end{itemize}

\begin{table*}[t]
\centering
\small
\setlength{\tabcolsep}{4pt}
\begin{tabular}{lll}
\toprule
\textbf{Paradigm} & \textbf{Decision interface} &  \textbf{Multimodal fusion} \\
\midrule
Generation-based LLM AES
& Decode and parse score text
& Prompt/decoding dependent \\

Pooling-head classifier
& Pool$(H)$ + $(K{+}1)$-way classification head
& Usually representation-level \\

\textbf{DLOM (ours)}
& \textbf{LM-head score-logit extraction on fixed score tokens}
& \textbf{Decision-level gated fusion} \\
\bottomrule
\end{tabular}
\caption{Positioning of our approach against common decision interfaces used in AES.}
\label{tab:positioning}
\end{table*}

\section{Related Work}

\subsection{Automated Essay Scoring}
Automated essay scoring (AES) has progressed from feature-engineered systems based on surface, syntactic, and discourse cues~\cite{31,32,18,5} 
to neural representation learning with CNN/RNN architectures and pre-trained language models~\cite{33,6}. 
Recent transformer-based models further improve scoring by leveraging contextual representations and transfer learning~\cite{34,35,36}, 
enabling end-to-end learning of essay representations and scoring functions.

\subsection{LLM-Based Essay Scoring}
With the rise of large language models (LLMs), many recent AES methods use instruction-following LLMs as backbones~\cite{37,38}. 
Most work adopts a generation-based interface, prompting the model to output a score token or short response aligned with rubric-defined grades~\cite{39,7}. 
This approach is simple and often requires no architectural changes, but the final label is typically obtained by decoding and parsing generated text, 
making it sensitive to prompting, decoding, and tokenization details. 
Several studies improve robustness via prompt design, calibration, or decoding constraints~\cite{43,42}, while largely retaining the generation paradigm.
\subsection{Ordinal Modeling for Scoring Tasks}
Ordinal modeling has been recognized as important for scoring and assessment tasks, where score categories follow a fixed order and misclassification costs are asymmetric~\cite{46,47}.
Prior work outside the LLM context has explored ordinal regression, ranking-based losses, and distance-aware objectives to better reflect the structure of scoring scales~\cite{48,49,50}. 
In LLM-based AES, however, ordinal structure is often only implicitly encoded through textual descriptions of score levels or prompt instructions~\cite{51}. 
Such an explicit decision interface over ordered labels has received relatively limited attention in LLM-based AES, motivating our decision-level ordinal formulation.

\paragraph{Positioning.}
Table~\ref{tab:positioning} summarizes common decision interfaces in LLM-based AES.
Compared to generation-and-parse scoring and pooling-head classifiers, DLOM reuses the LM head to extract score-wise logits on fixed score tokens and makes predictions directly in the ordered score space, avoiding both decoding-time artifacts and an additional classification head.

\subsection{Multimodal Essay Scoring}
Beyond text-only AES, recent work has explored multimodal essay scoring scenarios in which essays are accompanied by visual inputs such as charts, tables, or diagrams~\cite{1}.
Most approaches perform multimodal fusion at the representation level, combining textual and visual information via concatenation, attention mechanisms, or cross-modal transformers~\cite{2}.
Since modality relevance can vary across essays and traits, representation-level fusion may be sensitive to heterogeneous modality contributions.
Decision-level fusion that operates on score predictions has been less explored in AES, and offers an alternative interface for controlling modality influence and improving interpretability.

\begin{figure*}[t]
    \centering
    \includegraphics[width=\textwidth]{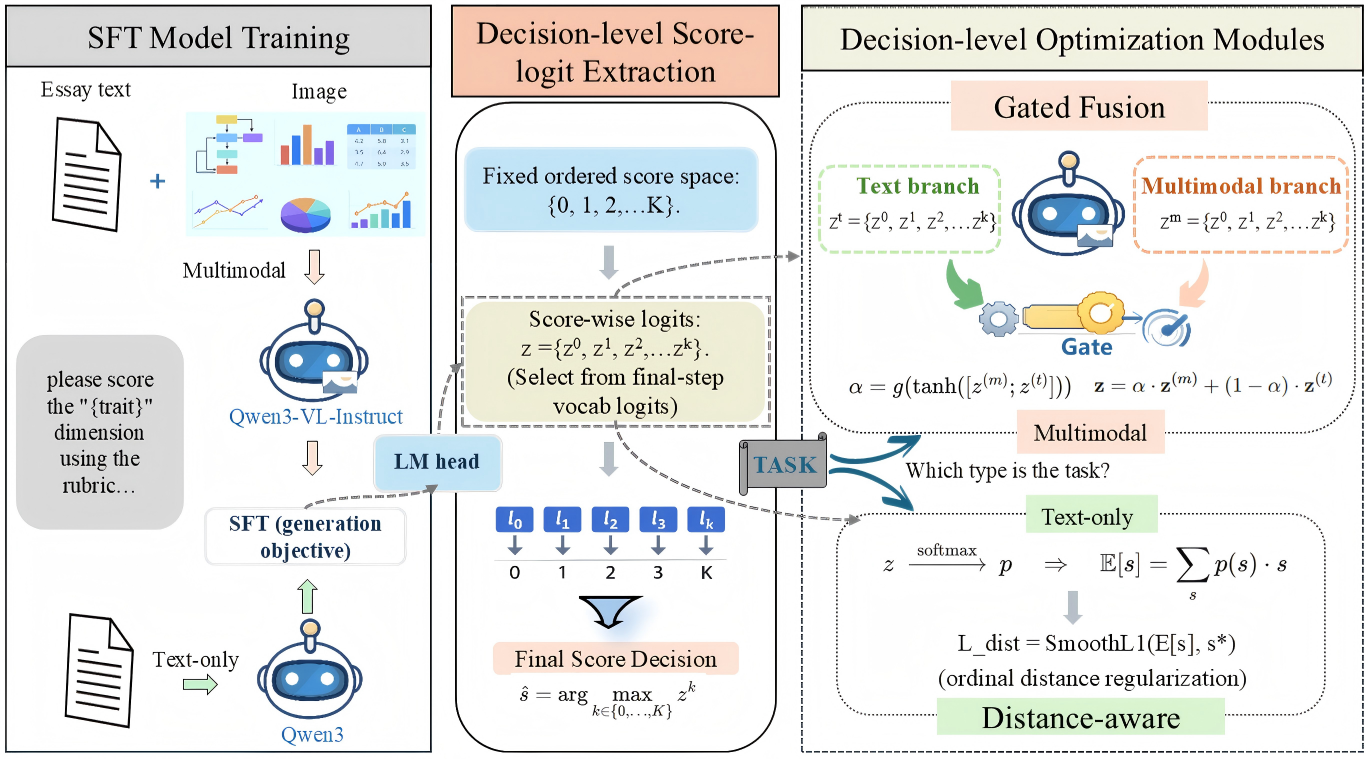}
    \caption{Overview of the proposed decision-level ordinal modeling framework.
    The framework consists of three stages: (i) supervised fine-tuning (SFT) for semantic encoding,
    (ii) decision-level score-logit extraction over an ordered score-token set,
    and (iii) task-specific decision-level objectives: decision-level gated fusion for multimodal scoring
    and distance-aware regularization for text-only scoring.}
    \label{fig:framework}
\end{figure*}

\section{Methodology}
\subsection{Problem Formulation}

Given an essay $e$ and an optional visual input $v$, the goal of automated essay scoring
is to predict a discrete score $y$ from a fixed, ordered set of score levels
$\mathcal{Y} = \{0, 1, \dots, K\}$.
The score levels follow an inherent ordinal structure, where misclassifications
between adjacent levels are less severe than those between distant levels. We assume access to a supervised fine-tuned (SFT) large language model that encodes
the semantic content of the essay.
Our objective is to design a decision mechanism that operates on top of this model and explicitly accounts for the ordinal nature of the scoring task at the decision stage.

\subsection{Decision-Level Ordinal Modeling}

Instead of formulating essay scoring as a language generation task,
we cast score prediction as an explicit decision over a fixed set of ordered score levels.
A standard AES classifier typically adds a $(K\!+\!1)$-way linear head on top of pooled hidden states (e.g., [CLS]) and predicts scores in that head space.
In contrast, our decision-level formulation reuses the LLM's LM head (vocabulary projection) and derives score-wise logits directly from the model outputs without decoding.
Concretely, we restrict the final-step vocabulary logits to a predefined set of score tokens, obtaining $\mathbf{z}\in\mathbb{R}^{K+1}$ as the decision space.
This keeps the backbone unchanged while making the scoring decision explicit in an ordered score space, and also supports decision-level fusion for multimodal scoring.

Figure~\ref{fig:framework} provides an overview of the proposed framework.
Given an input essay (and optional visual input), the SFT LLM produces final-step vocabulary logits via its LM head.
We select the entries corresponding to the predefined score tokens to form the score-wise logit vector
$\mathbf{z} = [z_0, z_1, \dots, z_K] \in \mathbb{R}^{K+1}$.
Formally, $\mathbf{z} = \mathrm{Logits}(e, v)_{L}[\mathcal{S}]$, where $\mathcal{S}$ denotes the score-token set and $L$ is the final position.
This formulation decouples semantic understanding from decision making:
the LLM encodes essay semantics, while the scoring decision is made explicitly in the ordinal score space.
At inference time, the predicted score is obtained by $\hat{y} = \arg\max_{k \in \mathcal{Y}} z_k$.
A detailed comparison between generation-based scoring, CLS-head classification, and our decision-level interface is provided in Appendix~\ref{appendix:decision_interface}.

\subsection{DLOM-GF for Multimodal AES}
For multimodal essay scoring settings where both textual and visual inputs are available, the reliability of visual information varies across essays and scoring traits. We introduce a decision-level gating mechanism that adaptively combines
textual and multimodal predictions based on their score-level evidence. Specifically, we obtain two score-wise logit vectors:
$\mathbf{z}^{(t)} \in \mathbb{R}^{K+1}$ from the text-only branch and
$\mathbf{z}^{(m)} \in \mathbb{R}^{K+1}$ from the multimodal branch,
where each logit corresponds to a candidate score level.
We concatenate the two logit vectors and apply a non-linear transformation
to produce a gating input:
\[
\mathbf{h} = \tanh([\mathbf{z}^{(m)};\mathbf{z}^{(t)}]).
\]
A lightweight gating network $g(\cdot)$ then maps $\mathbf{h}$ to a scalar
fusion weight:
\[
\alpha = g(\mathbf{h}) \in [0,1],
\]
where $g(\cdot)$ is implemented as a linear layer followed by a sigmoid activation.
The final score logits are computed as a convex combination in the ordinal score space:
\[
\mathbf{z} = \alpha \cdot \mathbf{z}^{(m)} + (1 - \alpha) \cdot \mathbf{z}^{(t)}.
\]

By conditioning the gate on score-wise logits rather than intermediate representations,
the proposed mechanism operates entirely at the decision level.
The gating function is learned implicitly through the scoring objective,
allowing the model to estimate modality reliability in an instance-adaptive
and task-aligned manner.

\subsection{Training Objective}

The model is trained to predict an ordinal score label from a fixed and ordered score set.
Given the score-wise logits $\mathbf{z}\in\mathbb{R}^{K+1}$ produced by the decision-level score-logit extraction
(or the gated fusion module in multimodal settings), we compute a categorical distribution
$p=\mathrm{softmax}(\mathbf{z})$ and optimize the cross-entropy loss with respect to the gold score $y$:
\[
\mathcal{L}_{\mathrm{CE}} = -\log p_{y}.
\]

For multimodal essay scoring, the score logits $\mathbf{z}$ are obtained via the proposed decision-level gated fusion module,
and the entire model is trained end-to-end by backpropagating $\mathcal{L}_{\mathrm{CE}}$ through both modality branches and the gating function.

In text-only settings, where decision-level fusion is not applicable, we additionally introduce a distance-aware regularization term
to better capture the ordinal structure of the scoring scale.
Specifically, we compute the expected score under the predicted distribution,
$\mathbb{E}[s] = \sum_{k\in\mathcal{Y}} p_k \cdot k$,
and penalize its deviation from the gold score using a SmoothL1 loss:
\[
\mathcal{L}_{\mathrm{dist}} = \mathrm{SmoothL1}(\mathbb{E}[s], y).
\]
The final objective is
\[
\mathcal{L} = \mathcal{L}_{\mathrm{CE}} + \lambda\,\mathcal{L}_{\mathrm{dist}} + \beta(\lambda-0.5)^2,
\]
where $\lambda=\sigma(\lambda_{\mathrm{logit}})\in(0,1)$ is a learnable weight and $\beta$ is a small regularization coefficient.
This auxiliary objective encourages ordinal consistency by assigning larger penalties to predictions that are farther away from the ground-truth score.

\begin{table*}[t]
\centering
\small
\setlength{\tabcolsep}{4pt}
\begin{tabular}{lccccccccccc}
\toprule
\multirow{2}{*}{Model} 
& \multicolumn{2}{c}{Lexical Level} 
& \multicolumn{4}{c}{Sentence Level} 
& \multicolumn{4}{c}{Discourse Level} 
& \multirow{2}{*}{Avg} \\
\cmidrule(lr){2-3} \cmidrule(lr){4-7} \cmidrule(lr){8-11}
& LA & LD & CH & GA & GD & PA & AC & JP & OS & EL &  \\
\midrule
CAFES
& 0.510 & 0.500 & 0.520& 0.570 & 0.540 & 0.490 & 0.370 & 0.440 & 0.480 & 0.280 & 0.470 \\
\midrule
SFT-Gen (baseline)
& 0.595 & 0.547 & \textbf{0.570} & 0.569 & 0.591 & 0.503 & 0.205 & 0.404 & 0.486 & 0.455 & 0.492 \\

DLOM
& 0.594 & 0.544 & 0.562 & 0.580 & 0.587 & 0.498 & \textbf{0.251} & 0.425 & 0.517 & 0.477 & 0.504 \\

DLOM-GF
& \textbf{0.624} & \textbf{0.551} & 0.569 & \textbf{0.589} & \textbf{0.613} & \textbf{0.506} & \textbf{0.251} & \textbf{0.447} & \textbf{0.527} & \textbf{0.479} & \textbf{0.516} \\
\bottomrule
\end{tabular}
\caption{Main results on the EssayJudge dataset under the multi-trait scoring setting.
LA: lexical accuracy, LD: lexical diversity, CH: coherence,
GA: grammatical accuracy, GD: grammatical diversity, PA: punctuation accuracy,
AC: argument clarity, JP: justifying persuasiveness,
OS: organizational structure, EL: essay length.
}
\label{tab:main_results}
\end{table*}
\section{Experiments}

\subsection{Datasets and Evaluation Metrics}

We evaluate our approach on both multimodal and text-only multi-trait essay scoring benchmarks and treat each trait as an independent ordinal prediction task and report trait-wise results as well as their macro-averaged performance.

\paragraph{Multimodal Dataset.}
For multimodal essay scoring, we use the EssayJudge dataset~\cite{1},
which consists of student essays accompanied by visual inputs such as charts or diagrams.
Each essay is annotated with multiple trait-level scores according to a predefined rubric,
covering diverse aspects of writing quality.
This dataset is specifically designed to assess multimodal reasoning in essay scoring,
where visual information may contribute unevenly across essays and traits (more information is in the Appendix~\ref{multimodal}).

\paragraph{Text-only Datasets.}
For text-only evaluation, we use the combined ASAP/ASAP++ benchmark, 
which is a standard dataset for multi-trait automated essay scoring.
The original ASAP dataset provides holistic scores for all prompts, while only a subset of prompts includes trait-level annotations.
ASAP++ extends ASAP by providing trait scores for the remaining prompts~\cite{9,8},
enabling comprehensive multi-trait evaluation across all prompts.
The final dataset consists of eight writing prompts, each associated with its own
scoring rubric and traits to be evaluated (more information is in the Appendix~\ref{text-only}).
Following prior work, we use the combined ASAP/ASAP++ dataset
for trait-level and cross-prompt evaluation.

\paragraph{Evaluation Metrics.}
We adopt Quadratic Weighted Kappa (QWK) as the primary evaluation metric~\cite{52},
which is widely used in AES to measure agreement between model predictions
and human raters while accounting for the ordinal nature of score scales. More information is shown in Appendix~\ref{appendix:qwk}.

\subsection{Experimental Setup}

\paragraph{Backbone Models.}
We use Qwen3-VL-8B-Instruct~\cite{10} as the backbone language model for multimodal experiments
and Qwen3-8B~\cite{11} for text-only experiments.
Both models are initialized from publicly released checkpoints and further
adapted to the essay scoring task through supervised fine-tuning. For further details on the model checkpoints, please refer to Appendix~\ref{appendix:models}.

\paragraph{Training Protocol.}
For both multimodal and text-only settings, we first perform generation-based supervised fine-tuning using a unified prompt template based on the official rubric descriptions. We then apply the proposed DLOM training on top of the SFT models, details of the prompt template and rubrics are provided in Appendix~\ref{appendix:training}.

\begin{table*}[t]
\centering
\small
\setlength{\tabcolsep}{4pt}
\begin{tabular}{lccccccccccc}
\toprule
\multirow{2}{*}{Model}
& \multicolumn{10}{c}{Traits} 
& \multirow{2}{*}{Avg} \\
\cmidrule(lr){2-11}
& Content & PA & Lang & Nar & Org & Conv & WC & SP & Style & Voice &  \\
\midrule
HISK
& 0.679 & 0.697 & 0.605 & 0.659 & 0.610 & 0.527 & 0.579 & 0.553 & 0.609 &0.489
& 0.601 \\
MTL-BiLSTM
& 0.685 & 0.701 & 0.604 & 0.668 & 0.615 & 0.560 & 0.615 & 0.598 & 0.632 &0.582
& 0.626 \\
ArTS
&0.730 & 0.751 & 0.698 & 0.725 & 0.672 & 0.668 & 0.679 & 0.678 & 0.721 &0.570
& 0.689 \\
\midrule
SFT-Gen (baseline)
& 0.705 & 0.752 & 0.698 & 0.732 & 0.638 & 0.582 & 0.632 & 0.614 & 0.521 & 0.585 & 0.646 \\

DLOM
& 0.738 & 0.776 & 0.716 & 0.753 & 0.692 & 0.655 & 0.659 & 0.662 & 0.615 & 0.581 & 0.685 \\

DLOM-DA
& \textbf{0.745} & \textbf{0.778} & \textbf{0.720} & \textbf{0.756} & \textbf{0.699} & \textbf{0.659} & \textbf{0.673} & \textbf{0.669} & \textbf{0.644} & \textbf{0.634} & \textbf{0.697} \\
\bottomrule
\end{tabular}
\caption{Trait-level results on the ASAP/ASAP++ benchmark.
PA: prompt adherence, Lang: language, Nar: narrativity,
Org: organization, Conv: conventions,
WC: word choice, SP: sentence fluency.
}
\label{tab:asappp_traits}
\end{table*}

In multimodal experiments, the text-only and multimodal branches are trained jointly, with score predictions from both branches combined through the decision-level gated fusion mechanism. The entire model, including the gating function, is optimized end-to-end under the same ordinal classification objective. All experiments are conducted using five-fold cross-validation, and results are reported by averaging performance across folds. For the text-only datasets (ASAP/ASAP++), we follow the previously established
partitioning scheme~\cite{27}.
For the multimodal dataset, which lacks an existing standard split, we construct
five folds at the essay level, ensuring that no essay appears in more than one fold.
We do not apply early stopping during training.
Instead, all models are trained for a fixed number of epochs under the same
training configuration.

\paragraph{Inference.}
At inference time, for both text-only and multimodal settings, scores are predicted directly from score-wise logits,
and each trait is evaluated independently.

\begin{table*}[t]
\centering
\small
\setlength{\tabcolsep}{5pt}
\begin{tabular}{lccccccccc}
\toprule
\multirow{2}{*}{Model}
& \multicolumn{8}{c}{Prompts} 
& \multirow{2}{*}{Avg}  \\
\cmidrule(lr){2-9}
& 1 & 2 & 3 & 4 & 5 & 6 & 7 & 8 &  \\
\midrule
HISK
& 0.674 & 0.586 & 0.651 & 0.681 & 0.693 & 0.709 & 0.641 & 0.516
& 0.644 \\
MTL-BiLSTM
& 0.670 & 0.611 & 0.647 & 0.708 & 0.704 & 0.712 & 0.684 & 0.581
& 0.665 \\
ArTS
& 0.708 & 0.706 & 0.704 & 0.767 & 0.723 & 0.776 & 0.749 & 0.603
& 0.717 \\
\midrule
SFT-Gen (baseline)
& 0.675 & 0.651 & 0.721 & 0.770 & 0.706 & 0.753 & 0.556 & 0.572 & 0.676 \\

DLOM
& \textbf{0.682} & \textbf{0.702} & 0.735 & 0.789 & \textbf{0.730} & 0.778 & 0.664 & 0.621 & 0.713 \\

DLOM-DA
& 0.681 & 0.696 & \textbf{0.743} & \textbf{0.795} & 0.728 & \textbf{0.780} & \textbf{0.688} & \textbf{0.646} & \textbf{0.720} \\
\bottomrule
\end{tabular}
\caption{Prompt-level results on the ASAP/ASAP++ benchmark.}
\label{tab:asap_prompts}
\end{table*}

\begin{figure*}[t]
    \centering
    \includegraphics[width=\textwidth]{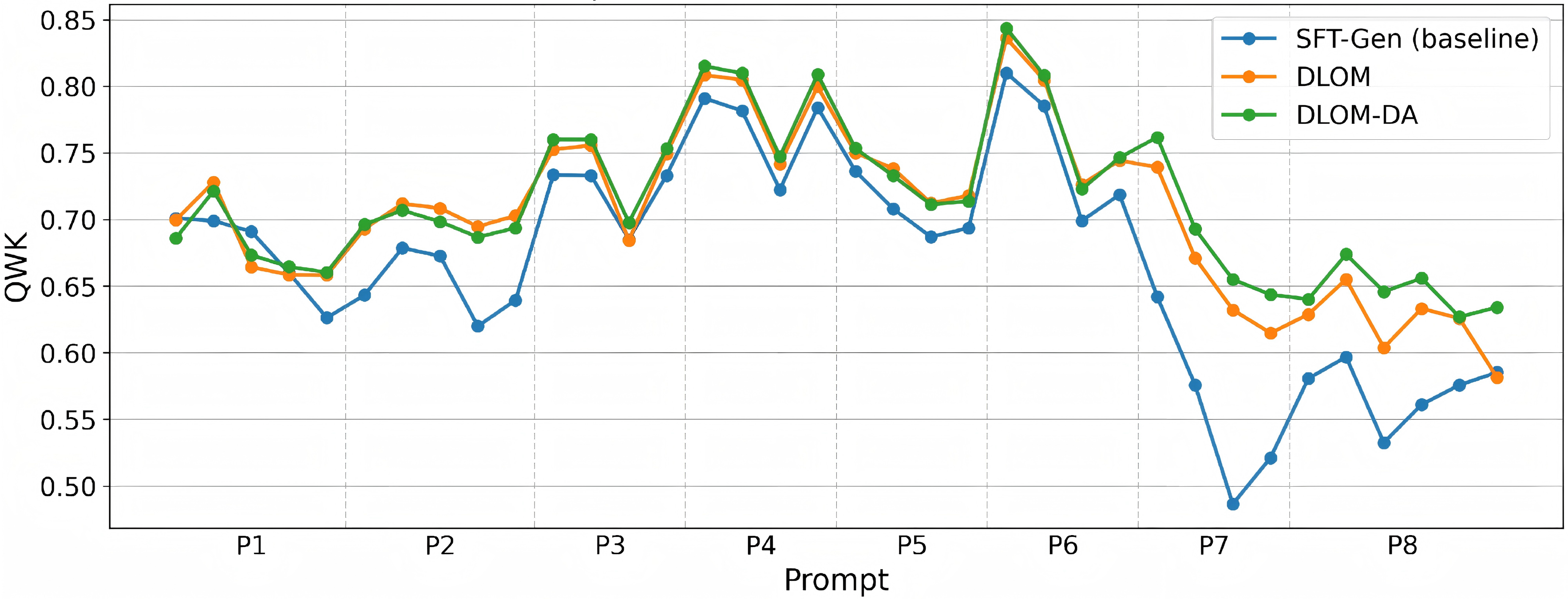}
    \caption{Prompt-wise QWK trends across different models on ASAP/ASAP++.
    Each point corresponds to a specific trait under a given prompt.
    Vertical dashed lines indicate boundaries between prompts.}
    \label{fig:qwk-mean-trend}
\end{figure*}
\subsection{Main Results on EssayJudge}
To enable supervised training and reproducible comparison on EssayJudge, we establish an explicit data partition and report training-based results. Prior work (CAFES) evaluates EssayJudge mainly through inference-time agentic pipelines over off-the-shelf MLLMs~\cite{2}. We therefore report their best-reported results in our table as a reference.
Table~\ref{tab:main_results} shows the main results on the multimodal EssayJudge
dataset under the multi-trait scoring setting.
We compare three scoring paradigms: generation-based supervised fine-tuning (SFT-Gen, baseline), decision-level ordinal modeling (DLOM), and decision-level modeling with gated multimodal fusion (DLOM-GF). 
Compared to SFT-Gen, decision-level ordinal modeling results in a higher
average QWK, while exhibiting differentiated effects across scoring traits.
In particular, the decision-level formulation yields clear gains on discourse-level
dimensions such as argument clarity and organizational structure, which require holistic
reasoning over essay content.
In contrast, improvements on lexical-level and sentence-level traits are comparatively
limited, suggesting that the benefits of explicit ordinal decision modeling are more
pronounced for higher-level scoring criteria.

Building upon the decision-level formulation, incorporating a gated multimodal fusion
mechanism further improves overall performance and achieves the best average QWK among
all compared methods.
Beyond improving individual lower-level traits, the gating mechanism consistently strengthens performance on discourse-level dimensions, further stabilizing gains on traits that rely on holistic essay evaluation. These results suggest that adaptively integrating textual and visual evidence at the decision stage not only alleviates weaknesses in specific traits, but also reinforces the advantages of decision-level modeling for high-level essay scoring. You can find more information about main results in Appendix~\ref{appendix:main results}

\subsection{Text-only Results on ASAP/ASAP++}

To examine whether the proposed decision-level formulation generalizes beyond
the multimodal setting, we further evaluate our approach on the text-only
ASAP/ASAP++ benchmark. For completeness, we also include several representative baselines (HISK~\cite{54}, MTL-BiLSTM~\cite{19}, and ArTS~\cite{3}). Overall, our variant DLOM-DA yields the best average performance among all compared methods.

Table~\ref{tab:asappp_traits} reports trait-level results on the dataset.
Compared to the baseline, our DLOM leads to
improvements on most traits and increases the average QWK from 0.646 to 0.685,
indicating the effectiveness of explicitly modeling ordinal score decisions. Incorporating distance-aware regularization further improves performance and
achieves the best average results among all compared methods, raising the average QWK to 0.697. Notably, the distance-aware variant yields consistent gains across all traits,
suggesting that the additional ordinal regularization helps stabilize
score predictions in the text-only setting.

The prompt-level results on the benchmark are in Table~\ref{tab:asap_prompts}.
The DLOM increases the average QWK from 0.676 to 0.713,
and incorporating distance-aware regularization further improves performance to 0.720. Specifically, the decision-level ordinal modeling achieves gains on all prompts. Overall, the text-only results corroborate our findings in the multimodal setting.
Even without visual inputs, explicitly modeling essay scoring at the decision level
leads to consistent performance improvements, demonstrating that the proposed
formulation's generalization ability. More details are provided in Appendix~\ref{appendix:auxiliary}.

\subsection{Analysis and Discussion}
\paragraph{Prompt-wise Analysis.}
To better understand the robustness of different modeling paradigms with respect to prompt variations, we conduct a prompt-wise analysis on the text-only dataset.
Figure~\ref{fig:qwk-mean-trend} presents QWK trends across prompts, where we evaluate the corresponding traits under each prompt.
While absolute performance varies across prompts and traits, SFT-Gen exhibits noticeable fluctuations, with a sustained performance degradation on P7 and P8 (i.e., consistently lower QWK across the associated traits). In contrast, DLOM yields more consistent trends across prompts, substantially mitigating extreme degradations observed in the SFT baseline.
This effect is further strengthened by incorporating distance-aware ordinal modeling, while DLOM-DA consistently maintains higher QWK scores across most prompts.
These observations suggest that explicitly modeling scoring decisions in the ordinal space reduces sensitivity to prompt-specific variations and leads to more robust essay scoring behavior.
\begin{table}[t]
\centering
\small
\setlength{\tabcolsep}{4pt}        
\renewcommand{\arraystretch}{0.95} 

\begin{tabular}{l c}
\toprule
Method & Avg. QWK \\
\midrule
SFT-Gen (baseline) & 0.492 \\
SFT + Ordinal Loss ($w{=}0.01$) & 0.472 \\
SFT + Ordinal Loss ($w{=}0.1$) & 0.481 \\
DLOM (ours) & \textbf{0.504} \\
\bottomrule
\end{tabular}

\vspace{-2mm}
\caption{Effect of decision-level formulation on the multimodal EssayJudge dataset.}
\label{tab:necessity}
\vspace{-2mm}
\end{table}

\begin{table*}[t]
\centering
\small
\setlength{\tabcolsep}{5pt}
\begin{tabular}{lccccccccccc}
\toprule
\textbf{Inference Strategy} 
& \textbf{LA} 
& \textbf{LD} 
& \textbf{CH} 
& \textbf{GA} 
& \textbf{GD} 
& \textbf{PA} 
& \textbf{AC} 
& \textbf{JP} 
& \textbf{OS} 
& \textbf{EL} 
& \textbf{Avg.} \\
\midrule
Text-only Decision 
& 0.524 & 0.551 & 0.474 & 0.541 & 0.559 & 0.414 & 0.249 & 0.342 & 0.525 & 0.363 & 0.454 \\
Multimodal-only Decision 
& 0.616 & 0.520 & 0.544 & 0.580 & 0.602 & 0.499 & 0.242 & 0.416 & 0.515 & 0.472 & 0.501 \\
DLOM-GF (ours) 
& \textbf{0.624} & \textbf{0.551} & \textbf{0.569} & \textbf{0.589} & \textbf{0.613} & \textbf{0.506} & \textbf{0.251} & \textbf{0.447} & \textbf{0.527} & \textbf{0.479} & \textbf{0.516} \\
\bottomrule
\end{tabular}
\caption{Per-trait comparison of different decision-level inference strategies on the multimodal EssayJudge dataset.}
\label{tab:per_trait_decision_fusion}
\end{table*}
\paragraph{Necessity of Decision-level Formulation.}

A natural question is whether ordinal structure can be incorporated into
generation-based essay scoring by simply adding an ordinal loss term,
without modifying the underlying modeling formulation.
In this setting, the ordinal loss is introduced as an auxiliary regularization term,
with $w$ controlling its relative weight to the token-level
generation objective.
To examine this alternative, we compare decision-level ordinal modeling
with generation-based SFT augmented with ordinal loss under different values of $w$ on the multimodal dataset.
As shown in Table~\ref{tab:necessity}, across the tested loss weights,
generation-based models with ordinal regularization consistently underperform
decision-level ordinal modeling, and even fall below the original
SFT baseline.

These results demonstrate that explicitly reformulating essay scoring as a decision over a fixed,
ordered score space leads to a higher average QWK.
Unlike loss-level ordinal constraints, which act as auxiliary signals on
token-level generation, decision-level ordinal modeling embeds ordinality
directly into the prediction space and the decision process itself.
This formulation encourages the model to perform score-wise comparisons
and reason explicitly over relative score preferences, rather than relying on
indirect token-level supervision.
These results suggest that, under our experimental setting,
ordinal structure in essay scoring is more effectively leveraged
when modeled at the decision level rather than imposed solely
through loss-level regularization.

\paragraph{Effect of Decision-Level Fusion.}
Building on the decision-level formulation, we further investigate whether fusing modality-specific score logits can yield additional gains in multimodal scoring.
Table~\ref{tab:per_trait_decision_fusion} presents a per-trait comparison of different
decision-level inference strategies on the EssayJudge dataset.
Overall, decision-level fusion achieves the highest average QWK (0.516),
outperforming both text-only (0.454) and multimodal-only (0.501) inference.
This indicates that, even when the backbone and the decision objective are fixed,
fusing modality-specific score logits at the decision stage provides a clear additional benefit.

At the trait level, decision-level fusion attains the best performance
across all scoring dimensions. 
While multimodal-only inference generally outperforms text-only inference, it still consistently lags behind decision-level fusion.
These results suggest that textual and visual cues provide complementary information,
and that the adaptive decision-level gating mechanism can selectively leverage the more informative modality,
leading to more reliable multimodal essay scoring decisions.

\paragraph{Design Implications.}
Our analysis suggests several practical insights for LLM-based essay scoring.
First, in the text-only setting, decision-level ordinal modeling exhibits improved robustness to prompt variations, and additional distance-aware regularization can further enhance performance stability.
Second, the key to leveraging ordinality lies in the decision-level formulation: by casting scoring as an explicit decision over a fixed ordered score set, ordinality is encoded directly in the prediction space and the decision mechanism. 
Third, for multimodal essay scoring, decision-level fusion with an adaptive gate provides an effective way to integrate textual and visual cues in the ordinal score space without requiring representation-level fusion.
Overall, these observations highlight the importance of aligning modeling choices with the structure of the scoring decision itself.

\section{Conclusion}
In this work, we revisited LLM-based automated essay scoring from a decision-centric perspective.
We reformulated scoring as an explicit ordinal decision over a fixed, ordered score set, enabling direct optimization in the score space without relying on decode-and-parse generation.
On top of this formulation, we introduced a decision-level gated fusion module for multimodal AES and a distance-aware regularization variant for text-only scoring.

Empirically, decision-level ordinal modeling improves over generation-based SFT on the multimodal EssayJudge dataset, and gated fusion yields further gains when visual relevance is heterogeneous.
On the text-only ASAP/ASAP++ benchmark, the same formulation remains effective, with distance-aware regularization providing additional improvements.
Overall, our results highlight that explicitly modeling the ordinal nature of scoring can yield more reliable LLM-based assessment, and the same view may extend to other ordinal or decision-centric evaluation tasks.

\section*{Limitations}
This work focuses on decision-level modeling for essay scoring and evaluates its effectiveness on a limited set of backbone models and datasets.
Although the proposed formulation and fusion strategy demonstrate consistent improvements, our experiments are conducted primarily with Qwen-based large language models.
Further validation on a broader range of architectures may be necessary to assess the generality of the approach.

In addition, while our decision-level fusion framework is applicable to multimodal settings, distance-aware regularization is only explored in the text-only scenario.
We leave a systematic investigation of integrating additional ordinal constraints with multimodal fusion to future work.

\bibliography{custom}
\clearpage
\appendix
\section{Additional Clarification on Decision Interfaces}
\label{appendix:decision_interface}

This appendix clarifies how our decision-level formulation differs from two common alternatives for LLM-based AES: generation-based scoring and standard classification heads.

\paragraph{Generation-based scoring.}
A generation-based scorer predicts a score by decoding a textual output (e.g., a score token or a short phrase) and then mapping the decoded text to an ordinal label.
Formally, it relies on $\hat{y}=\mathrm{Parse}(\mathrm{Decode}(p_\theta(\cdot \mid e,v)))$,
which introduces sensitivity to decoding strategies (e.g., sampling/greedy) and prompt/output formatting.

\paragraph{CLS-head classification (with/without ordinal losses).}
A standard classifier treats the LLM as an encoder, pools hidden states, and applies an additional $(K{+}1)$-way head:
\[
\mathbf{o}=W_{\text{cls}}\cdot \mathrm{pool}(H) + b,\quad \hat{y}=\arg\max_k o_k,
\]
optionally augmented with ordinal regression objectives (e.g., CORAL-style losses or distance-based regularizers) on top of $\mathbf{o}$.
This design introduces a separate decision space parameterized by $W_{\text{cls}}$, which is not tied to the LM output distribution.

\paragraph{Ours: decision-level score-logit extraction.}
In contrast, we reuse the LM head and make the decision directly in a predefined ordered score space by restricting the final-step vocabulary logits to a fixed score-token set $\mathcal{S}$:
\[
\mathbf{z}=\mathrm{Logits}(e,v)_{L}[\mathcal{S}] \in \mathbb{R}^{K+1},\quad \hat{y}=\arg\max_{k\in\mathcal{Y}} z_k.
\]
This removes the need for decoding and avoids introducing an additional classification head, while yielding an explicit score-wise decision representation.
Operating in this score-logit space also enables our decision-level gated fusion, which adaptively combines modality-specific evidence via a convex combination of $\mathbf{z}^{(m)}$ and $\mathbf{z}^{(t)}$ under the same decision objective.
Moreover, this interface is LLM-native: it leverages the model’s learned token-preference distribution and avoids relying on prompt- or decoding-specific behaviors when producing a final score.

\section{Experiments}
\label{appendix:A}
\subsection{EssayJudge}
\label{multimodal}
Table~\ref{essayjudge} shows the detailed information about EssayJudge dataset~\cite{1}. It consists of 1,054 multimodal essays written as part of IELTS Writing Task 1. In this task, candidates are required to write a report based on visual data such as charts, graphs, or diagrams. The dataset is specifically designed to assess multimodal reasoning in automated essay scoring, where visual inputs play a significant role in the essay's content and quality. 
\begin{table}[htbp]
  \centering
  \begin{tabular}{lc}
    \toprule
    \textbf{Statistic} & \textbf{Number} \\
    \midrule
    Total Multimodal Essays & 1,054 \\
    \midrule
    Image Type & \\
    \hspace{0.5em}- Single-Image & 703 (66.7\%) \\
    \hspace{0.5em}- Multi-Image & 351 (33.3\%) \\
    \midrule
    Multimodal Essay Type & \\
    \hspace{0.5em}- Flow Chart & 305 (28.9\%) \\
    \hspace{0.5em}- Bar Chart & 211 (20.0\%) \\
    \hspace{0.5em}- Table & 153 (14.5\%) \\
    \hspace{0.5em}- Line Chart & 145 (13.8\%) \\
    \hspace{0.5em}- Pie Chart & 71 (6.7\%) \\
    \hspace{0.5em}- Map & 62 (5.9\%) \\
    \hspace{0.5em}- Composite Chart & 107 (10.2\%) \\
    \bottomrule
  \end{tabular}
  \caption{Key statistics of EssayJudge dataset.}
  \label{essayjudge}
\end{table}

\subsection{ASAP/ASAP++}
\label{text-only}
We use the open-sourced \texttt{ASAP/ASAP++}, as summarized in Table~\ref{asap/asap++}, different prompts are assessed with distinct traits, each having varied score ranges~\cite{9,8}. It consists of English essays written by American 7th to 10th-grade high school students across eight prompts. 
\begin{table}[htbp]
  \centering
  \resizebox{\linewidth}{!}{ 
    \begin{tabular}{l|c|c|c}
      \toprule
      Prompt & \# Essays & Traits & Score Range \\ 
      \midrule
      1 & 1785 & Content, WC, Org, SF, Conv & 1 - 6\\
      2 & 1800 & Content, WC, Org, SF, Conv & 1 - 6\\
      3 & 1726 & Content, PA, Nar, Lang & 0 - 3\\
      4 & 1772 & Content, PA, Nar, Lang & 0 - 3\\
      5 & 1805 & Content, PA, Nar, Lang & 0 - 4\\
      6 & 1800 & Content, PA, Nar, Lang & 0 - 4\\
      7 & 1569 & Content, Org, Conv, Style & 0 - 6\\
      8 & 723  & Content, WC, Org, SF, Conv, Voice & 2 - 12\\
      \bottomrule
    \end{tabular}
  }
  \caption{Composition of the ASAP/ASAP++ combined dataset. The prompt defines the writing theme. Abbreviations: WC (Word Choice), Org (Organization), SF (Sentence Fluency), Conv (Conventions), PA (Prompt Adherence), Nar (Narrativity), Lang (Language).}
  \label{asap/asap++}
\end{table}

\begin{table*}[t]
  \centering
  \begin{tabular}{|l|l|l|}
    \hline
    \textbf{Model} & \textbf{Source} & \textbf{URL} \\
    \hline
    Qwen3-VL-8B-Instruct & local checkpoint & \url{https://huggingface.co/Qwen/Qwen3-VL-8B-Instruct} \\
    \hline
    Qwen3-8B & local checkpoint & \url{https://huggingface.co/Qwen/Qwen3-8B} \\
    \hline
  \end{tabular}
  \caption{Model sources for the multimodal and text-only essay scoring tasks.}
  \label{tab:model_checkpoints}
\end{table*}
\subsection{Quadratic Weighted Kappa (QWK)}
\label{appendix:qwk}
Quadratic Weighted Kappa (QWK) is a commonly used metric in automated essay scoring (AES) to measure the agreement between model predictions and human raters. It accounts for the ordinal nature of the score scales, where errors between adjacent score levels are less severe than those between distant levels. 

The QWK is defined as:

\[
QWK = 1 - \frac{\sum_{i,j} w_{ij} \cdot O_{ij}}{\sum_{i,j} w_{ij} \cdot E_{ij}}
\]

where:
- \( O_{ij} \) is the observed agreement between categories \(i\) and \(j\),
- \( E_{ij} \) is the expected agreement between categories \(i\) and \(j\),
- \( w_{ij} \) is the weight assigned to the difference between score levels \(i\) and \(j\).

The weight \( w_{ij} \) is typically defined as:

\[
w_{ij} = \frac{(i - j)^2}{(K - 1)^2}
\]

where \( K \) is the number of possible score categories.

\subsection{Model Selecting and  Sources}
\label{appendix:models}
For our experiments, we selected two large language models: Qwen3-VL-8B-Instruct and Qwen3-8B, chosen for their distinct strengths in multimodal and text-only tasks, respectively.

\paragraph{\textbf{Qwen3-VL-8B-Instruct}}: This is a multimodal model designed to handle both textual and visual inputs, making it ideal for essay scoring tasks where essays are accompanied by images, charts, or other forms of visual data. And its open-source availability makes it accessible for replication and future work in this area.

\paragraph{\textbf{Qwen3-8B}}: This is a text-only model, suitable for tasks that involve only textual input. We use it as a baseline to compare the performance of our multimodal model, allowing us to isolate the effect of incorporating visual information into the scoring process.

These two models represent a balanced comparison between multimodal and text-only settings, allowing us to evaluate the robustness and generality of our proposed method across different types of inputs. Table~\ref{tab:model_checkpoints} shows their checkpoints' source.

\subsection{Prompt Template and Rubrics}
\label{appendix:training}
\begin{figure}[htbp]
    \centering
    \includegraphics[width=\linewidth]{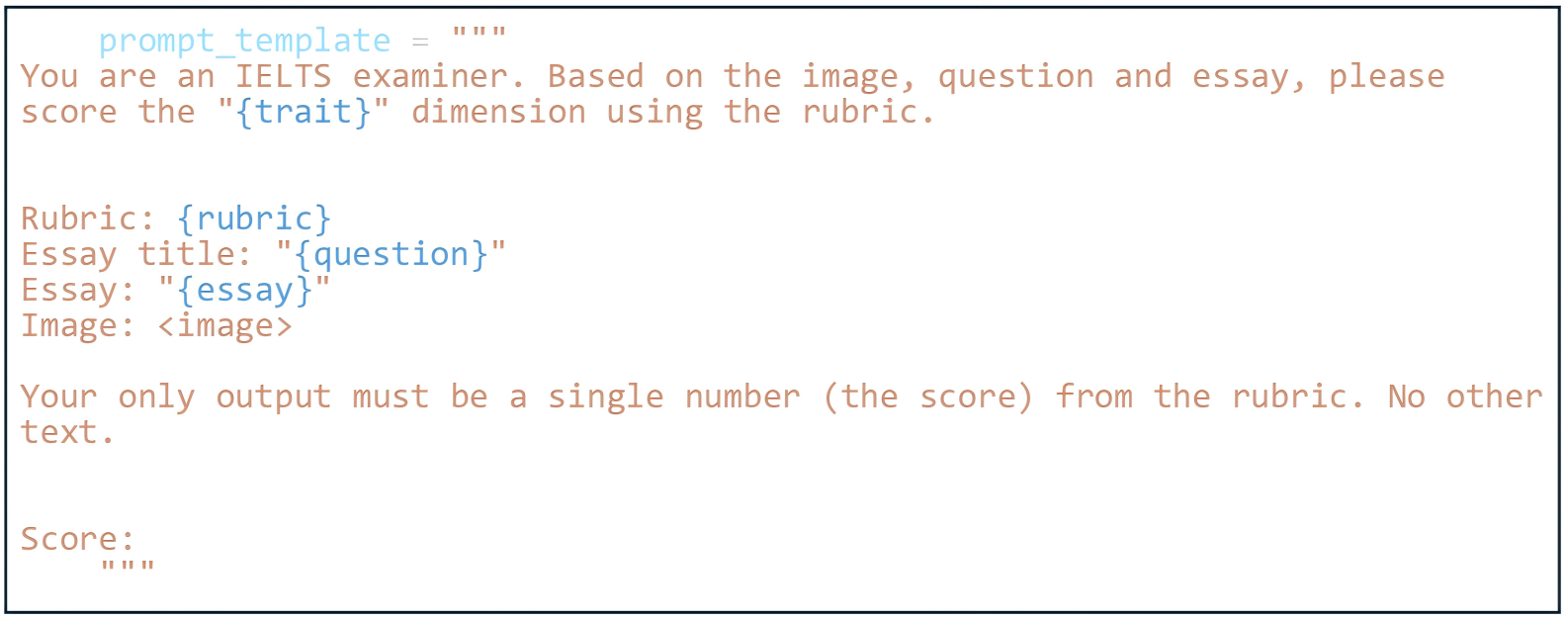}
    \caption{Prompt Template for Multimodal Dataset}
    \label{fig:template-1}
\end{figure}

\paragraph{Training on multimodal dataset.} The training was conducted using a unified prompt template as shown in Figure~\ref{fig:template-1} and the rubric for each trait is publicly available and can be accessed in the original dataset repository~\cite{1}.

\begin{figure}[htbp]
    \centering
    \includegraphics[width=\linewidth]{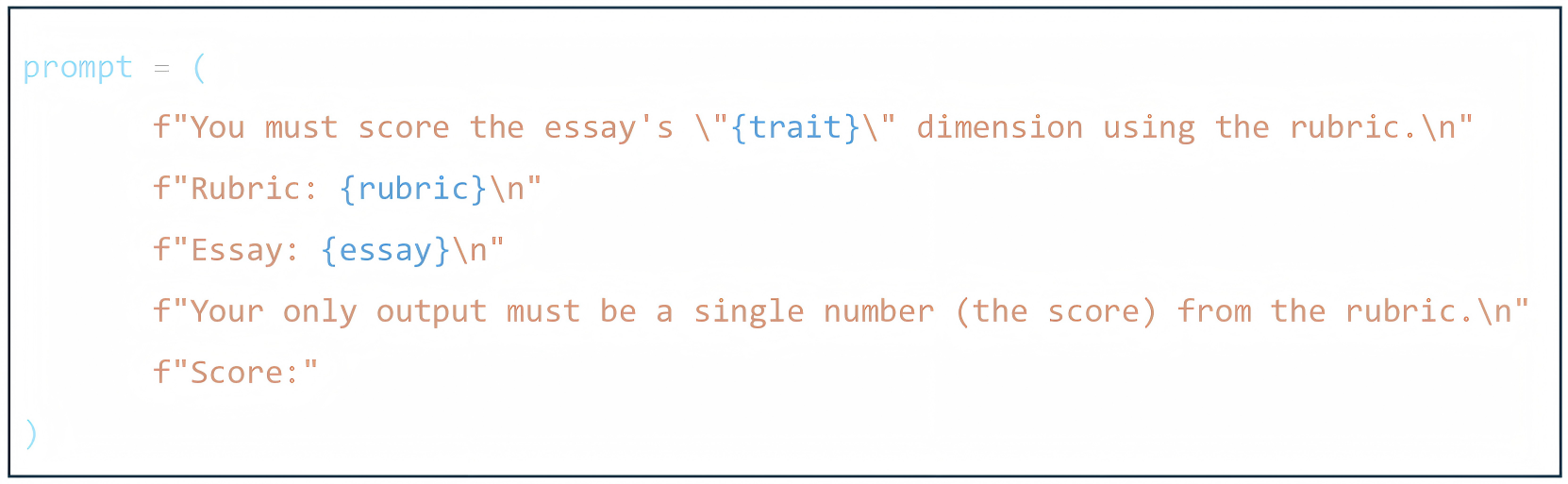}
    \caption{Prompt Template for Text-only Dataset}
    \label{fig:template-2}
\end{figure}

\paragraph{Training on text-only datasets.}
We train all text-only models using a unified prompt template (Figure~\ref{fig:template-2}).
Trait rubrics and prompt definitions for ASAP/ASAP++ are publicly available.\footnote{
ASAP: \url{https://www.kaggle.com/c/asap-aes}. \;
ASAP++: \url{https://lwsam.github.io/ASAP++/Irec2018.html}.
}

\begin{table*}[htbp]
    \centering
    \small
    \setlength{\tabcolsep}{1.5pt}
    \begin{tabular}{lccccccccccc}
        \toprule
        Model & Content & PA & Lang & Nar & Org & Conv & WC & SP & Style & Voice & Avg \\
        \midrule
        SFT (Generation)
        & $\pm0.031$ & $\pm0.026$ & $\pm0.026$ & $\pm0.025$ & $\pm0.051$ & $\pm0.066$ & $\pm0.044$ & $\pm0.049$ & $\pm0.030$ & $\pm0.056$ & $\pm0.040$ \\
        Decision-level (Ordinal)
        & $\pm0.026$ & $\pm0.025$ & $\pm0.028$ & $\pm0.025$ & $\pm0.036$ & $\pm0.036$ & $\pm0.044$ & $\pm0.032$ & $\pm0.028$ & $\pm0.083$ & $\pm0.036$ \\
        Decision-level + Distance-aware
        & $\pm0.030$ & $\pm0.023$ & $\pm0.035$ & $\pm0.031$ & $\pm0.035$ & $\pm0.038$ & $\pm0.037$ & $\pm0.032$ & $\pm0.020$ & $\pm0.083$ & $\pm0.036$ \\
        \bottomrule
    \end{tabular}
    \caption{Standard deviation of QWK across cross-validation folds on the ASAP/ASAP++ benchmark, reported at the trait level.}
    \label{tab:asap_trait_std}
\end{table*}

\begin{table*}[htbp]
    \centering
    \small
    \setlength{\tabcolsep}{4pt}
    \begin{tabular}{lccccccccc}
        \toprule
        Model & 1 & 2 & 3 & 4 & 5 & 6 & 7 & 8 & Avg \\
        \midrule
        SFT (Generation) 
        & $\pm0.053$ & $\pm0.036$ & $\pm0.032$ & $\pm0.028$ & $\pm0.021$ & $\pm0.017$ & $\pm0.054$ & $\pm0.057$ & $\pm0.037$ \\
        Decision-level (Ordinal)
        & $\pm0.020$ & $\pm0.033$ & $\pm0.033$ & $\pm0.018$ & $\pm0.035$ & $\pm0.016$ & $\pm0.034$ & $\pm0.054$ & $\pm0.031$ \\
        Decision-level + Distance-aware
        & $\pm0.029$ & $\pm0.031$ & $\pm0.040$ & $\pm0.020$ & $\pm0.036$ & $\pm0.017$ & $\pm0.030$ & $\pm0.054$ & $\pm0.032$ \\
        \bottomrule
    \end{tabular}
    \caption{Standard deviation of QWK across cross-validation folds on the ASAP/ASAP++ benchmark, reported at the prompt level.}
    \label{tab:asap_prompt_std}
\end{table*}
\subsection{Main Results}
\label{appendix:main results}
We observe relatively large variance across cross-validation folds for several traits,
including the generation-based baseline.
This variance is mainly attributed to the limited size and heterogeneous nature of the
EssayJudge dataset, rather than instability introduced by the proposed methods.
Importantly, decision-level modeling and gated fusion do not exhibit higher variance than
the baseline, indicating comparable stability under the same experimental setting.

\begin{figure}[t]
    \centering
    \includegraphics[width=\linewidth]{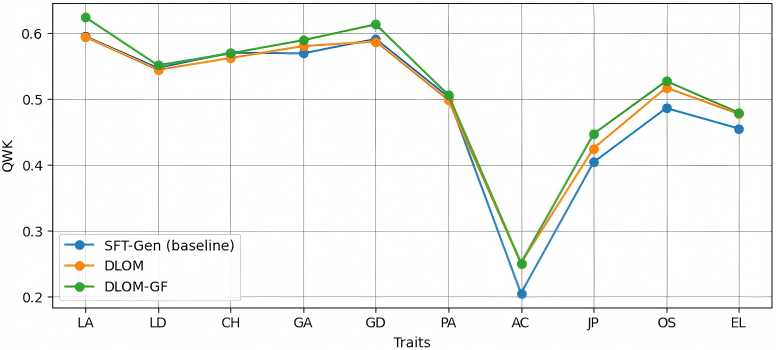}
    \caption{Trait-wise QWK comparison on the EssayJudge dataset.
    }
    \label{fig:essayjudge_traits}
\end{figure}

Figure~\ref{fig:essayjudge_traits} provides a visualization of the trait-level QWK results
reported in Table~\ref{tab:main_results}.
The figure highlights consistent improvements of decision-level modeling over the
generation-based baseline across most traits, with more pronounced gains on
discourse-level dimensions.
The gated fusion mechanism further strengthens performance, particularly on traits
with lower baseline scores.

\subsection{Text-only Results on ASAP/ASAP++}
\label{appendix:auxiliary}

\begin{figure}[t]
    \centering
    \includegraphics[width=\linewidth]{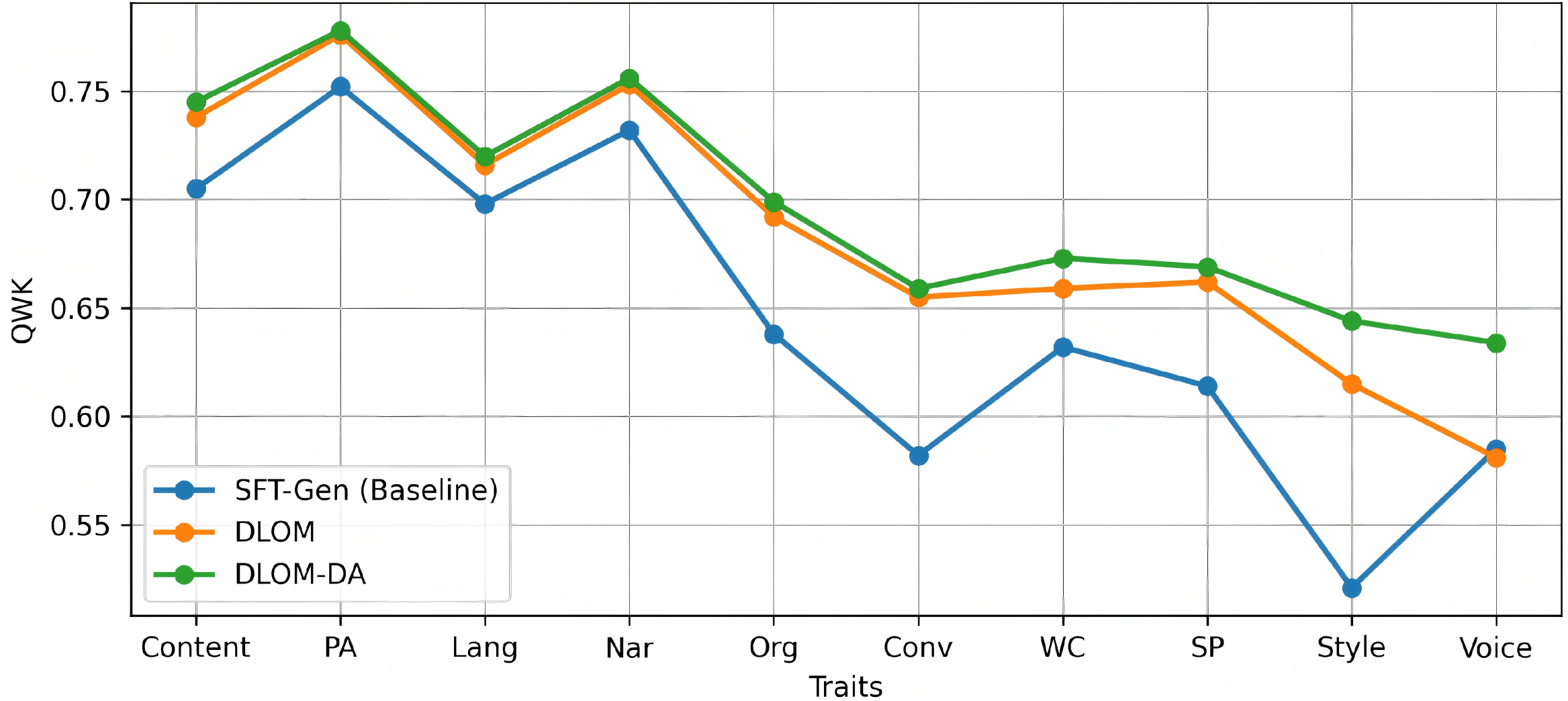}
    \caption{Trait-wise QWK trends on the ASAP/ASAP++ benchmark.
    }
    \label{fig:asap_traits}
\end{figure}

\begin{figure}[t]
    \centering
    \includegraphics[width=\linewidth]{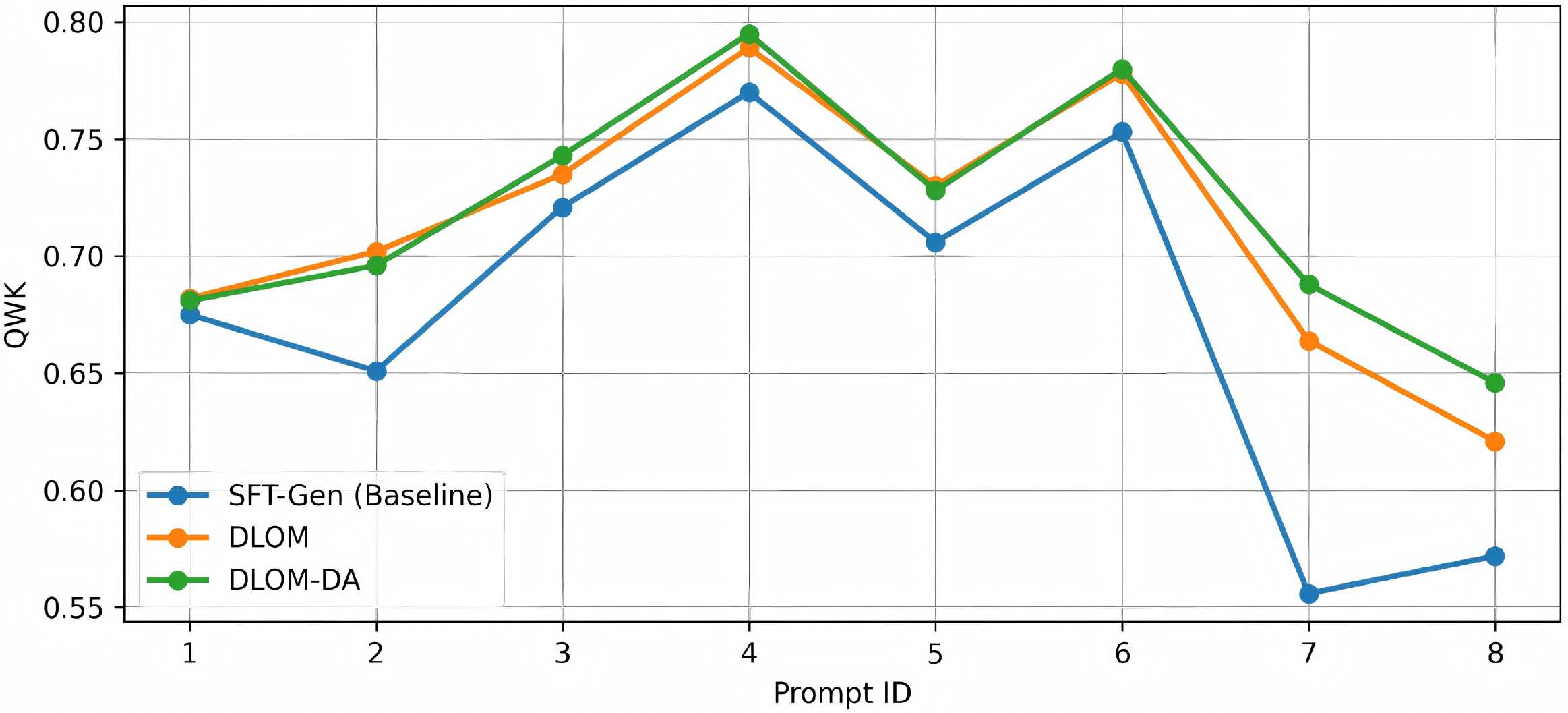}
    \caption{Prompt-wise QWK trends on the ASAP/ASAP++ benchmark.
   }
    \label{fig:asap_prompts}
\end{figure}

\paragraph{Variance analysis.}
Tables~\ref{tab:asap_trait_std} and~\ref{tab:asap_prompt_std} report the standard deviation of QWK across cross-validation folds
at the trait and prompt levels.
Across both views, decision-level modeling and the distance-aware variant achieve higher mean performance without increasing variability:
the average standard deviation remains comparable to the generation baseline.
This suggests that the observed improvements are not obtained at the cost of reduced stability.
\paragraph{Visualizations of text-only results.}
Figures~\ref{fig:asap_traits} and~\ref{fig:asap_prompts} visualize the trait-level and prompt-level
QWK results from Tables~\ref{tab:asappp_traits} and~\ref{tab:asap_prompts}, respectively.
Overall, decision-level modeling improves over the generation-based baseline on most traits and prompts,
and the distance-aware variant provides further (often more consistent) gains.
Improvements are particularly pronounced on higher-level traits that require holistic judgments of content and organization.

\end{document}